\documentclass[sigconf]{acmart}
\usepackage{subcaption}
\usepackage{booktabs} 
\usepackage[english]{babel}
\usepackage[export]{adjustbox}
\usepackage{mathrsfs}
\usepackage{csquotes}
\usepackage{multirow}
\usepackage{multicol}
\usepackage{array}
\usepackage{arydshln}
\usepackage{booktabs}
\usepackage{siunitx}
\usepackage[inline]{enumitem}
\usepackage{xspace}
\usepackage{etoolbox}
\usepackage{scalerel}

\makeatletter
\patchcmd\@combinedblfloats{\box\@outputbox}{\unvbox\@outputbox}{}{%
   \errmessage{\noexpand\@combinedblfloats could not be patched}%
}%
 \makeatother
 
\newlist{inlinelist}{enumerate*}{1}
\setlist*[inlinelist,1]{%
  label=(\roman*),
}
\author{Mohammad Aliannejadi}
\affiliation{%
  \institution{Universit{\`a} della Svizzera italiana (USI)}
}
\email{mohammad.alian.nejadi@usi.ch}

\author{Hamed Zamani}
\affiliation{%
  \institution{University of Massachusetts Amherst}
}
\email{zamani@cs.umass.edu}

\author{Fabio Crestani}
\affiliation{%
  \institution{Universit{\`a} della Svizzera italiana (USI)}
}
\email{fabio.crestani@usi.ch}

\author{W. Bruce Croft}
\affiliation{%
  \institution{University of Massachusetts Amherst}
}
\email{croft@cs.umass.edu}

\newcommand{\dataname}{Qulac\xspace}
\newcommand{\modelname}{NeuQS\xspace}
\newcommand{\qrmodelname}{BERT-LeaQuR\xspace}

\newcommand{\partitle}[1]{\vspace{1mm}\noindent\textbf{#1}}

\begin{document}

\fancyhead{}
\title[]{Asking Clarifying Questions in Open-Domain Information-Seeking Conversations}

\begin{abstract}

Users often fail to formulate their complex information needs in a single query. As a consequence, they may need to scan multiple result pages or reformulate their queries, which may be a frustrating experience.
Alternatively, systems can improve user satisfaction by proactively asking questions of the users to \textit{clarify} their information needs. Asking clarifying questions is especially important in conversational systems since they can only return a limited number of (often only one) result(s).

In this paper, we formulate the task of asking clarifying questions in open-domain information-seeking conversational systems. To this end, we propose an offline evaluation methodology for the task and collect a dataset, called  \textit{\dataname}, through crowdsourcing. Our dataset is built on top of the TREC Web Track 2009-2012 data and consists of over 10K question-answer pairs for 198 TREC topics with 762 facets.
Our experiments on an oracle model demonstrate that asking only one good question leads to over $170\%$ retrieval performance improvement in terms of P@1, which clearly demonstrates the potential impact of the task. We further propose a retrieval framework consisting of three components: question retrieval, question selection, and document retrieval. In particular, our question selection model takes into account the original query and previous question-answer interactions while selecting the next question. Our model significantly outperforms competitive baselines. To foster research in this area, we have made \dataname publicly available.

\end{abstract}

\copyrightyear{2019} 
\acmYear{2019} 
\setcopyright{acmcopyright}
\acmConference[SIGIR '19]{Proceedings of the 42nd International ACM SIGIR Conference on Research and Development in Information Retrieval}{July 21--25, 2019}{Paris, France}
\acmBooktitle{Proceedings of the 42nd International ACM SIGIR Conference on Research and Development in Information Retrieval (SIGIR '19), July 21--25, 2019, Paris, France}
\acmPrice{15.00}
\acmDOI{10.1145/3331184.3331265}
\acmISBN{978-1-4503-6172-9/19/07}

\maketitle

\vspace{-0.3cm}
\section{Introduction}
\label{sec:intro}

\begin{figure}
    \centering
    \includegraphics[width=\columnwidth]{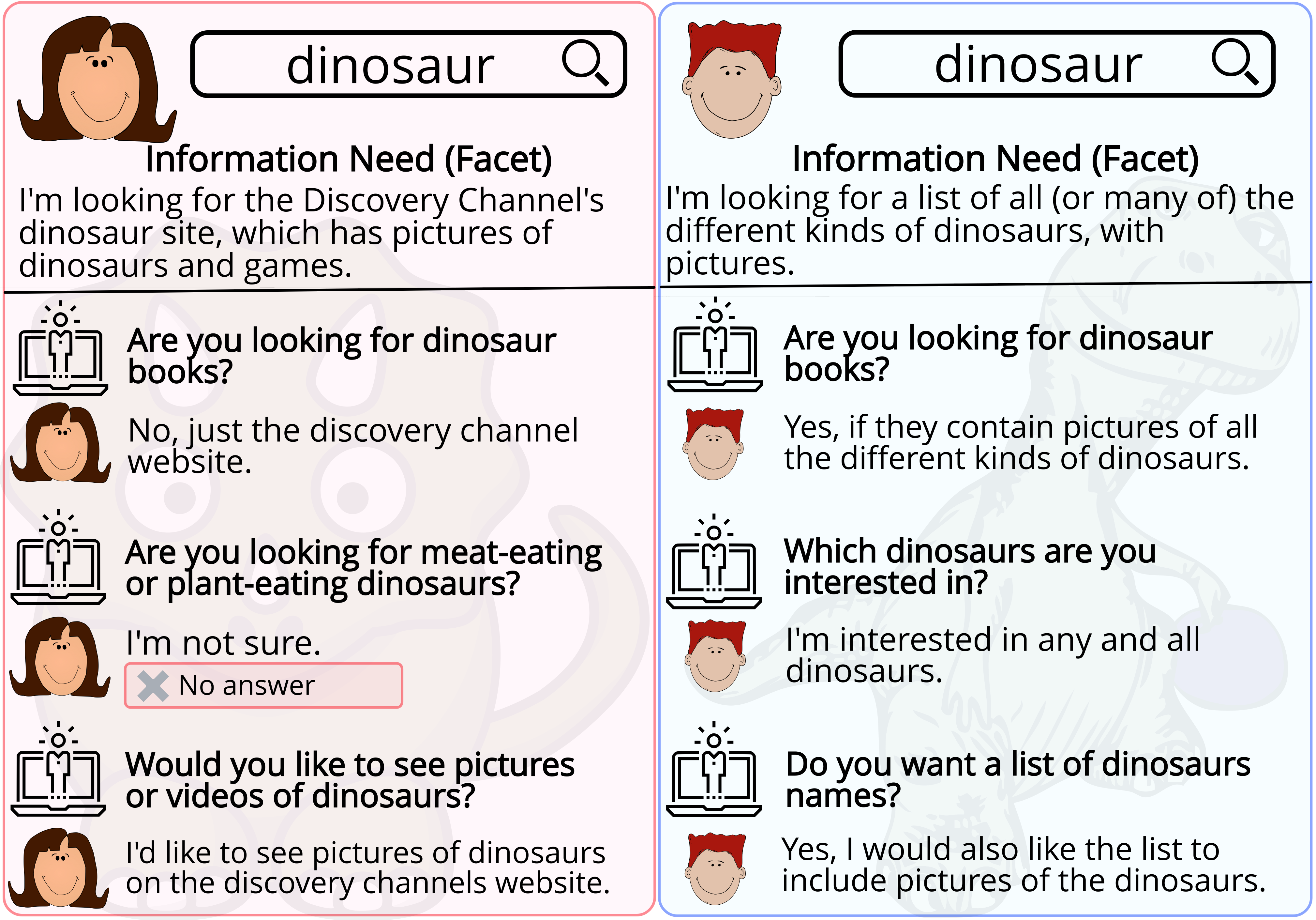}
    \vspace{-0.7cm}
    \caption{Example conversations with clarifying questions from our dataset, Qulac. As we see, both users, Alice and Robin, issue the same query (``dinosaur''), however, their actual information needs are completely different. With no prior knowledge, the system starts with the same clarifying question. Depending on the user's answers, the system selects the next questions in order to clarify the user's information need. The tag ``No answer'' shows that the asked question is not related to the information need.
    }
    \vspace{-0.05cm}
    \label{fig:exmaple}
\end{figure}

While searching on the Web, users often fail to formulate their complex information needs in a single query. As a consequence, they may need to scan multiple result pages or reformulate their queries. 
Alternatively, systems can decide to proactively ask questions to clarify users' intent before returning the result list~\cite{DBLP:conf/chiir/BraslavskiSAD17, DBLP:conf/chiir/RadlinskiC17}. In other words, a system can assess the level of confidence in the results and decide whether to return the results or ask questions from the users to \emph{clarify} their information need. 
The questions can be aimed to clarify ambiguous, faceted or incomplete queries~\cite{DBLP:conf/chi/VtyurinaSAC17}. 
Asking clarifying questions is especially important in conversational search systems for two reasons:
    \begin{inlinelist}
        \item conversation is the most convenient way for natural language interactions and asking questions~\cite{DBLP:conf/sigir/KieselBSAH18} and
        \item a conversational system can only return a limited number of results, thus being confident about the retrieval performance becomes even more important.
    \end{inlinelist}  Asking clarifying questions is a possible solution for improving this confidence. Figure \ref{fig:exmaple} shows an example of such a conversation selected from our dataset. We see that both users, Alice and Robin, issue the same query, ``dinosaur.'' Assuming that the system does not have access to any prior personal or contextual information, the conversation starts with the same clarifying question. The rest of the conversation, however, depends on the users' responses. In fact, the users' responses aid the system to get a better understanding of the underlying information need. 

\begin{figure}
    \centering
    \includegraphics[width=\columnwidth]{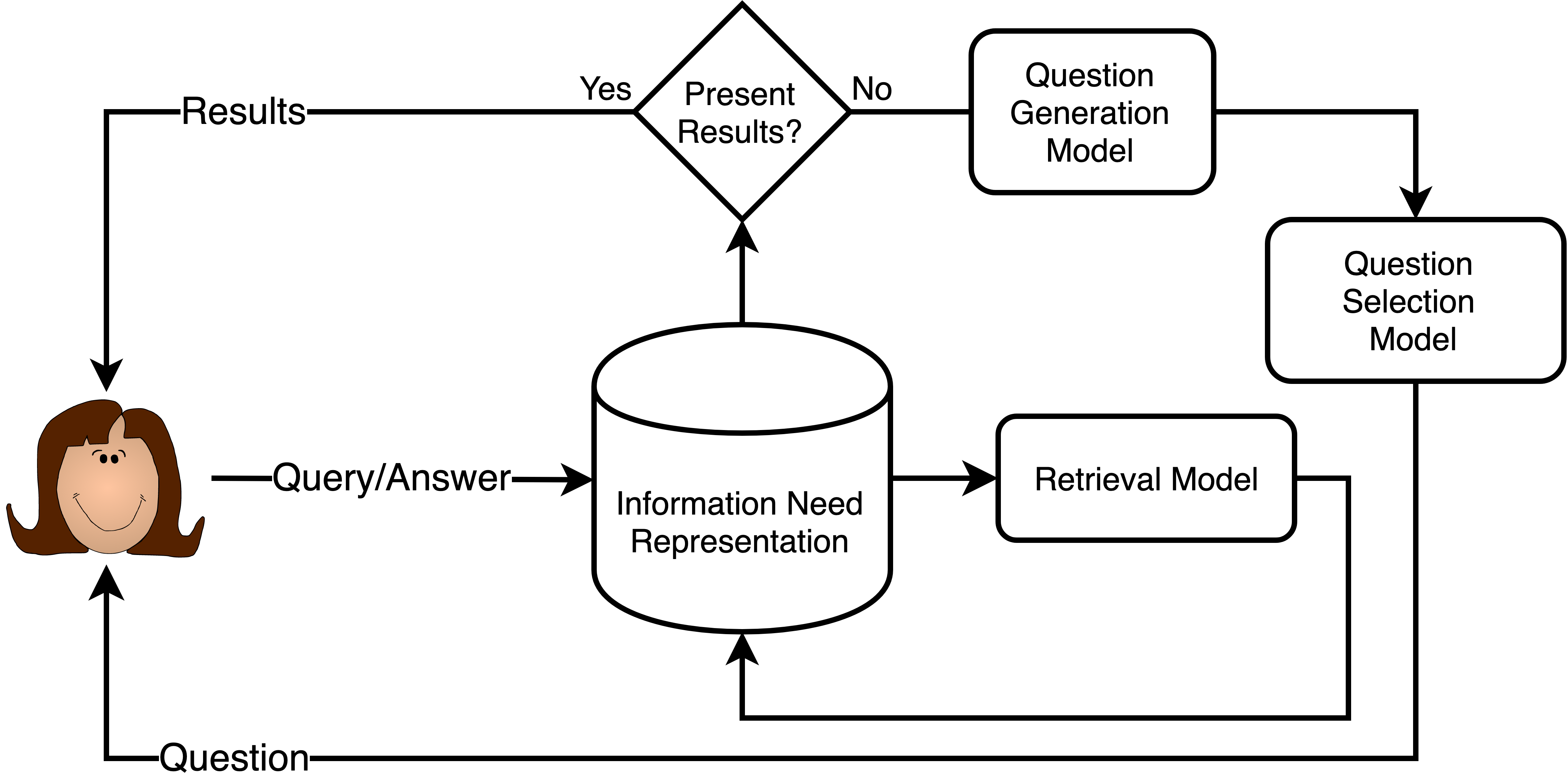}
    \vspace{-0.8cm}
    \caption{A workflow for asking clarifying questions in an open-domain conversational search system.}
    \vspace{+0.1cm}
    \label{fig:flowchart}
\end{figure}

A possible workflow for an information system with clarifying questions is shown in Figure \ref{fig:flowchart}. As we can see, Alice initiates a conversation by submitting her query to the system. The system then retrieves a list of documents and estimates its confidence on the result list (i.e., ``Present Results?''). If the system is not sufficiently confident to present the results to the user, it then starts the process of asking clarifying questions. As the first step, it generates a list of candidate questions related to Alice's query. Next, the system selects a question from the candidate question list and asks it from the user. Based on Alice's answer, the system retrieves new documents and repeats the process.

In this paper, we formulate the task of selecting and asking clarifying questions in open-domain information-seeking conversational systems. To this end, we propose an offline evaluation framework based on faceted and ambiguous queries and collect a novel dataset, called \textit{\dataname},\footnote{Qulac
means \textit{blizzard} and \textit{wonderful} in Persian.} building on top of the TREC Web Track 2009-2012 collections. \dataname consists of over 10K question-answer pairs for 198 TREC topics consisting of 762 facets. Inspired from successful examples of crowdsourced collections~\cite{Aliannejadi:2018,DBLP:conf/sigir/AlonsoS14}, we collected clarifying questions and their corresponding answers for every topic-facet pair via crowdsourcing. 
Our offline evaluation protocol enables further research on the topic of asking clarifying questions in a conversational search session, providing a benchmarking methodology to the community.

Our experiments on an oracle model show that asking only one good question leads to over $100\%$ retrieval performance improvement. 
Moreover, the analysis of the oracle model provides important intuitions related to this task. For instance, we see that asking clarifying questions can improve the performance of shorter queries more. Also, clarifying questions exhibit a more significant effect on improving the performance of ambiguous queries, compared to faceted queries.
We further propose a retrieval framework following the workflow of Figure~\ref{fig:flowchart}, consisting of three main components as follows: \begin{inlinelist}
    \item question retrieval; 
    \item question selection; and
    \item document retrieval.
\end{inlinelist} The question selection model is a simple yet effective neural model that takes into account both users' queries and the conversation context. We compare the question retrieval and selection models with competitive term-matching and learning to rank (LTR) baselines, showing their ability to significantly outperform the baselines.
Finally, to foster research in this area, we have made \dataname publicly available.\footnote{Code and data are available at \url{https://github.com/aliannejadi/qulac}.}

\vspace{-0.2cm}
\section{Related Work}
\label{sec:rel}

While conversational search has roots in early Information Retrieval (IR) research, the recent advances in automatic voice recognition and conversational agents have created increasing interest in this area.
One of the first works in conversational IR dates back to 1987 when \citet{DBLP:journals/jasis/CroftT87} proposed I$^3$R that acted as an expert intermediary system, communicating with the user in a search session. A few years later \citet{belkin1995cases} characterized information-seeking strategies for conversational IR, offering users choices in a search session based on case-based reasoning.
Since then researchers in the fields of IR and natural language processing (NLP) have studied various aspects of this problem. Early works focused on rule-base conversational systems~\cite{DBLP:conf/acl/WalkerPB01,DBLP:conf/sigdial/WilliamsRRB13}, while another line of research has investigated spoken language understanding approaches~\cite{225939,DBLP:journals/csl/HeY05,DBLP:conf/acl-alta/AliannejadiKKG14} for intelligent dialogue agents in the domain of flight~\cite{DBLP:conf/naacl/HemphillGD90} and train trip information~\cite{DBLP:journals/speech/AustOSS95}. The challenge was to understand the user's request and query a database of flight or train schedule information accordingly.
The recent advances of conversational agents have attracted research in various aspects of conversational information access~\cite{AliannejadiSigir18, DBLP:conf/sigir/SunZ18,DBLP:conf/sigir/YanSW16,DBLP:conf/chiir/BahrainianC18}.
One line of research analyzes data to understand how users interact with voice-only systems~\cite{DBLP:journals/jasis/SpinaTCS17}. \citet{DBLP:conf/chiir/RadlinskiC17} proposed a theoretical framework for conversational search highlighting the need for multi-turn interactions with users for narrowing down their specific information needs. Also, \citet{DBLP:conf/chiir/TrippasSCJS18} studied conversations of real users to identify the commonly-used interactions and inform a conversational search system design.
Moreover, research on query suggestion is relevant to our work if we consider suggesting queries as a means of clarifying users' intent in a traditional IR setting~\cite{DBLP:conf/chiir/RadlinskiC17}.
Result diversification and personalizing is one of the key components for query suggestion~\cite{DBLP:journals/kbs/JiangLYN15a}, especially when applied to small-screen devices. In particular, \citet{DBLP:conf/wsdm/KatoT16} found that presenting results for one facet and suggesting queries for other facets is more effective on such devices.

Research on clarifying questions has attracted considerable attention in the fields of NLP and IR.
People have studied human-generated dialogues on question answering (QA) websites, analyzing the intent of each utterance~\cite{DBLP:conf/sigir/QuYCTZQ18} and, more specifically, clarifying questions~\cite{DBLP:conf/chiir/BraslavskiSAD17}. \citet{DBLP:conf/sigir/KieselBSAH18} studied the impact of voice query clarification on user satisfaction and found that users like to be prompted for clarification.
Much work has been done on interacting with users for recommendation. 
For instance, \citet{DBLP:conf/kdd/Christakopoulou16} designed a system that can interact with users to collect more detailed information about their preferences in venue recommendation. 
Also, \citet{DBLP:conf/sigir/SunZ18} utilized a semi-structured user query with facet-value pairs to represent a conversation history and proposed a deep reinforcement learning framework to build a personalized conversational recommender system. 
Focusing on clarifying questions, \citet{DBLP:conf/cikm/ZhangCA0C18} automatically extracted facet-value pairs from product reviews and considered them as questions and answers. They proposed a multi-memory network to ask questions for improved e-commerce recommendation. 
Our work is distinguished from these studies by formulating the problem of asking clarifying questions in an open-domain information-seeking conversational setting where several challenges regarding extracting topic facets~\cite{DBLP:conf/sigir/KongA13} are different from a recommendation setting.

In the field of NLP, researchers have worked on question ranking~\cite{DBLP:conf/acl/DaumeR18} and generation~\cite{DBLP:journals/corr/abs-1904-02281,DBLP:conf/acl/HuangNWL18} for conversation. These studies rely on large amount of data from industrial chatbots~\cite{DBLP:conf/acl/HuangNWL18,DBLP:conf/acl/QiuYJZHCCL18}, query logs~\cite{DBLP:conf/www/RenNMK18}, and QA websites~\cite{DBLP:conf/acl/DaumeR18,DBLP:journals/corr/abs-1904-02281,DBLP:conf/acl/TianYMSFZ17}. For instance, \citet{DBLP:conf/acl/DaumeR18} proposed a neural model for question selection on a simulated dataset of clarifying questions and answers extracted from QA websites such as StackOverflow. Later, they proposed an adversarial training for generating clarifying questions for a given product description on Amazon~\cite{DBLP:journals/corr/abs-1904-02281}. Also, \citet{DBLP:conf/acl/HuangNWL18} studied the task of question generation for an industrial chatbot.
Unlike these works, we study the task of asking clarification question in an IR setting where the user's request is in the form of short queries (vs. a long detailed post on StackOverflow) and the system should return a ranked list of documents.

\vspace{-0.2cm}
\section{Problem Statement}
\label{sec:problem}

A key advantage of a conversational search system is its ability to interact with the user in the form question and answer.

In particular, a conversational search system can proactively pose questions to the users to understand their actual information needs more accurately and improve its confidence in the search results. We illustrate the workflow of a conversational search system, focusing on asking clarifying questions. 

As depicted in Figure~\ref{fig:flowchart}, once the user submits a query to the system, the \textit{Information Need Representation} module generates and passes their information need to the Retrieval Model, which returns a ranked list of documents. The system should then measure its confidence in the retrieved documents (i.e., \textit{Present Results?} in Figure~\ref{fig:flowchart}). In cases where the system is not sufficiently confident about the quality of the result list, it passes the query and the context (including the results list) to the \textit{Question Generation Model} to generate a set of clarifying questions, followed by the \textit{Question Selection Model} whose aim is to select one of the generated questions to be presented to the user. Next, the user answers the question and the same procedure repeats until a stopping criterion is met. Note that when the user answers a question, the complete session information is considered for selecting the next question. In some cases, a system can decide to present some results, followed by asking a question. For example, assume a user submits the query ``sigir 2019'' and the system responds ``The deadline of SIGIR 2019 is Jan. 28. Would you like to know where it will be held?'' As we can see, while the system is able to return an answer with high confidence, it can still ask further questions~\cite{DBLP:conf/sigir/YanZE17}. In this work, we do not study this scenario; however, one can investigate it for exploratory search.

\vspace{-0.35cm}
\subsection{A Facet-Based Offline Evaluation Protocol}
\label{sec:eval}

The design of an offline evaluation protocol is challenging because conversation requires online interaction between a user and a system. Hence, an offline evaluation strategy requires human-generated answers to all possible questions that a system would ask, something that is impossible to achieve in an offline setting. To circumvent this problem, we substitute the Question Generation Model in Figure \ref{fig:flowchart} with a large bank of questions, assuming that it consists of all possible questions in the collection.
Although this assumption is not absolutely realistic, it reduces the complexity of the evaluation significantly as human-generated answers to a limited set of questions can be collected offline, facilitating offline evaluation. 

In this work, we build our evaluation protocol on top of the TREC Web track's data. TREC has released 200 search topics, each of which being either ``ambiguous'' or ``faceted.''\footnote{In this work, we use the term ``facet'' to refer to the subtopics of both faceted and ambiguous topics.} \citet {DBLP:conf/trec/ClarkeCS09} defined these categories as follows:
``... Ambiguous queries are those that have multiple distinct interpretations.
... On the other hand, facets reflect underspecified queries, with different aspects covered by the subtopics...'' The TREC collection is originally designed to evaluate search result diversification. In contrast, here we build various conversation scenarios based on topic facets.

Formally, let $\mathcal{T} = \{t_1, t_2, \dots, t_n\}$ be the set of topics (queries) that initiates a conversation. Moreover, we define $\mathcal{F}=\{\mathbf{f}_1, \mathbf{f}_2, \dots,$ $ \mathbf{f}_n\}$ as the set of facets with $\mathbf{f}_i = \{f^i_{1}, f^i_{2}, \dots,  f^i_{m_{i}}\}$ defining different facets of $t_i$, where $m_{i}$ denotes the number of facets for $t_i$. Further, let $\mathcal{Q} = \{\mathbf{q}_1, \mathbf{q}_2, \dots, \mathbf{q}_n\}$ be the set of clarifying questions belonging to every topic, where $\mathbf{q}_i = \{q^i_1, q^i_2, \dots, q^i_{z_i}\}$ consists of all clarifying questions that belong to $t_i$; $z_i$ is the number of clarifying questions for $t_i$. 
Here, our aim is to provide the users' answers to all clarifying questions considering all topics and their corresponding facets. Therefore, let $\mathcal{A}(t, f, q) \rightarrow a$ define a function that returns answer $a$ for a given topic $t$, facet $f$, and question $q$. Hence, to enable offline evaluation, $\mathcal{A}$ requires to return an answer for all possible values of $t$, $f$, and $q$. In this work, 
$\mathcal{T}$ and $\mathcal{F}$ are borrowed from the TREC Web track 2009-2012 data. $\mathcal{Q}$ is then collected via crowdsourcing and $\mathcal{A}(t,f,q)$ is also modeled by crowdsourcing (see Section \ref{sec:col}). It is worth noting that we also borrow the relevance assessments of the TREC Web track, after breaking them down to the facet level. For instance, suppose the topic ``dinosaur'' has 10 relevant documents, 6 of which are labeled as relevant to the first facet, and 4 to the second facet. In \dataname, the topic ``dinosaur'' is broken into two topic-facet pairs together with their respective relevance judgments.

\vspace{-0.2cm}
\section{Data Collection}
\label{sec:col}

In this section, we explain how we collected \dataname (\textbf{Qu}estions for \textbf{la}ck of \textbf{c}larity), that is, to the best of our knowledge, the first dataset of clarifying questions in an IR setting. 
As we see in Figure \ref{fig:exmaple}, each topic is coupled with a facet. Therefore, the same question would receive a different answer based on the user's actual information need. We follow a four-step strategy to build \dataname. In the first step we define the topics and their corresponding facets. 
In the second step, we collect a number of candidate clarifying questions ($\mathcal{Q}$) for each query through crowdsourcing. 
Then, in the third step, we assess the relevance of the questions to each facet and collect new questions for those facets that require more specific questions.
Finally, in the last step, we collect the answers for every query-facet-question triplet, modeling $\mathcal{A}$. In the following subsections, we elaborate on every step of our data collection procedure. 

\vspace{-0.3cm}
\subsection{Topics and Facets}
As we discussed earlier, the problem of asking clarifying questions is particularly interesting in cases where a query can be interpreted in various ways. 
An example is shown in Figure \ref{fig:exmaple} where two different users issue the same query for different intents. Therefore, any data collection should contain an initial query and description of its facet, describing the user's information need. In other words, we define a target facet for each query. 
Faceted and ambiguous queries make an ideal case to study the effect of clarifying questions in a conversational search system for the following reasons:
\begin{inlinelist}
\item the user information need is not clear from the query;
\item multiple facets of the same query could satisfy the user's information need;
\item asking clarifying questions related to any of the facets provide a high information gain.
\end{inlinelist}
Therefore, we choose the TREC Web track's topics\footnote{\url{https://trec.nist.gov/data/webmain.html}}~\cite{DBLP:conf/trec/ClarkeCV12} as the basis for \dataname. In other words, we take the topics of TREC Web track 09-12 as initial user queries. We then break each topic into its facets and assume that each facet describes the information need of a different user (i.e., it is a topic).
As we see in Table \ref{tab:stats}, the average facet per topic is 3.85 $\pm$ 1.05. Therefore, the initial 198 TREC topics\footnote{
The official TREC relevance judgements cover 198 of the topics.
} leads to 762 topic-facet pairs in \dataname. Consequently, for each topic-facet pair, we take the relevance judgements associated with the respective facet. 

\vspace{-0.2cm}
\subsection{Clarifying Questions}
It is crucial to collect a set of reasonable questions that address multiple facets of every topic\footnote{Candidate clarifying questions should also address out-of-collection facets.} while containing sufficient negative samples. This enables us to study the effect of retrieval models under the assumption of having a functional question generation model. Therefore, we asked human annotators to generate questions for a given query based on the results they observed on a commercial search engine as well as query auto-complete suggestions. 

 To collect clarifying questions, we designed a Human Intelligence Task (HIT) on Amazon Mechanical Turk.\footnote{\url{http://www.mturk.com}} We asked the workers to imagine themselves acting as a conversational agent such as Microsoft Cortana where an imaginary user had asked them about a topic. Then, we described the concept of facet to them, supporting it with multiple examples. Finally, we asked them to follow the steps below to figure out the facets of each query and generate questions accordingly:
 
 \begin{enumerate}[leftmargin=*]
    \item Enter the same query in a search engine of their choice and scan the results in the first three pages. Reading the title of the results as well as scanning the snippets would give them an idea of different facets of the query on the Web.
    \item For some difficult queries such as ``toilet,'' scanning the results would not help in identifying the facets. Therefore, 
    inspired by \cite{DBLP:conf/wsdm/BenetkaBN17}, 
    we asked the workers to type the query in the search box of the search engine, and press the space key after typing the query. Most commercial search engines provide a list of query auto-complete suggestions.
    Interestingly, in most cases the suggested queries reflect various aspects of the same query.
    \item Finally, we asked them to generate six questions related to the query, aiming to address the facets that they had figured out.
\end{enumerate}

We assigned two workers to each HIT, resulting in 12 questions per topic in the first round. In order to preserve language diversity of the questions, we limited each worker to a maximum of two HITs. HITs were available to workers residing in the U.S. with an approval rate of over 97\%. 
After collecting the clarifying questions, in the next step, we explain how we verified them for quality assurance.

\vspace{-0.2cm}
\subsection{Question Verification and Addition}
\label{sec:verification}
In this step, we aim to address two main concerns: \begin{inlinelist}
\item how good are the collected clarifying questions?
\item are all facets addressed by at least one clarifying question?
\end{inlinelist}
Given the high complexity of this step, we appointed two expert annotators for this task. We instructed the annotators to read all the collected questions of each topic, marking invalid and duplicate questions. Moreover, we asked them to match a question to a facet if the question was \textit{relevant} to the facet. A question was considered relevant to a facet if its answer would address the facet. Finally, in order to make sure that all facets were covered by at least one question, we asked the annotators to generate an additional question for the facets that needed more specific questions. The outcome of this step is a set of verified clarifying questions, addressing all the facets in the collection.

\vspace{-0.2cm}
\subsection{Answers}
After collecting and verifying the questions, we designed another HIT in which we collected answers to the questions for every facet. The HIT started with detailed instructions of the task, followed by several examples. The workers were provided with a topic and a facet description. Then we instructed them to assume that they had submitted the query with their actual information need being the given facet.
Then they were required to write the answer to \textit{one} clarifying question that was presented to them.
To avoid the bias of other questions for the same facet, we included only one question in each HIT.
If a question required information other than what workers were provided with, we instructed the workers to identify it with a ``No answer'' tag.
Each worker was allowed to complete a maximum of 100 HITs to guarantee language diversity. Workers were based in the U.S. with an approval rate of 95\% or greater.

\partitle{Quality check.} During the course of data collection, we performed regular quality checks on the collected answers. The checks were done manually on 10\% of submissions per worker. In case we observed any invalid submissions among the sampled answers of one user, we then studied all the submissions of the same user. Invalid submissions were then removed from the collection and the worker was banned from the future HITs. Finally, we assigned all invalid answers to other workers to complete. Moreover, we employed basic behavioral check techniques in the design of the HIT. For example, we disabled copy/paste features of text inputs and tracked workers' keystrokes. This enabled us to detect and reject low-quality submissions.

\begin{table}[]
    \centering
    \caption{Statistics of \dataname.}
    \vspace{-0.3cm}
    \begin{tabular}{ll}
        \toprule
         \# topics & 198 \\
         \# faceted topics & 141\\
         \# ambiguous topics & 57\\
         \midrule
         \# facets & 762\\
         Average facet per topic & 3.85 $\pm$ 1.05\\
         Median facet per topic & 4 \\
         \# informational facets & 577\\
         \# navigational facets & 185\\
         \midrule
         \# questions & 2,639\\
         \# question-answer pairs & 10,277\\
         Average terms per question & 9.49 $\pm$ 2.53\\
         Average terms per answer & 8.21 $\pm$ 4.42\\
         \bottomrule
    \end{tabular}
    \label{tab:stats}
    \vspace{-0.5cm}
\end{table}

\vspace{-0.2cm}
\section{Selecting Clarifying Questions}
\label{sec:method}

In this section, we propose a conversational search system that is able to select and ask clarifying questions and rank documents based on the user's responses. The proposed system retrieves a set of questions for a given query from a large pool of questions, containing all the questions in the collection. At the second stage, our proposed model, called \modelname, aims to select the best question to be posed to the user based on the query and the conversation context. This problem is particularly challenging because the conversational interactions are in natural language, highly depending on the previous interactions between the user and the system (i.e., conversation context). 

As mentioned earlier in Section~\ref{sec:problem}, a user initiates the conversation by submitting a query. Then the system should decide whether to ask a clarifying question or present the results. At every stage of the conversation, the previous questions and answers exchanged between the user and the system are known to the model. Finally, the selected question and its corresponding answer should be incorporated in the document retrieval model to enhance the retrieval performance. 

Formally, for a given topic $t$ let $\mathbf{h} = \{(q_1,a_1), (q_2,a_2), \dots, (q_{|\mathbf{h}|}, $ $a_{|\mathbf{h}|})\}$ be the history of clarifying questions and their corresponding answers exchanged between the user and the system (i.e., context). Here, the ultimate goal is to predict $q$, that is the next question that the system should ask from the user. Moreover, let $a$ be the user's answer to $q$. The answer $a$ is unknown to the question selection model, however, the document retrieval model retrieves documents once the system receives the answer $a$. In the following, we describe the question retrieval model, followed by the question selection and the document retrieval models.

\vspace{-0.2cm}
\subsection{Question Retrieval Model}
\label{sec:qr}

We now describe our \textbf{BERT}\footnote{BERT: Bidirectional Encoder Representations from Transformers} \textbf{L}anguage R\textbf{e}present\textbf{a}tion based \textbf{Qu}estion \textbf{R}etrieval model, called \qrmodelname. We aim to maximize the recall of the retrieved questions, retrieving all relevant clarifying questions to a given query in the top $k$ questions. 
Retrieving all relevant questions from a large pool of questions is challenging, because questions are short and context-dependent. In other words, many questions depend on the conversation context and the query. Also, since conversation is in the form of natural language, term-matching models cannot effectively retrieve short  questions. For instance, some relevant clarifying questions for the query ``dinosaur'' are: ``Are you looking for a specific web page?'' ``Would you like to see some pictures?'' 

\citet{DBLP:journals/corr/YangZZGC17} showed that neural models outperform term-matching models for question retrieval. Inspired by their work, we learn a high-dimensional language representation for the query and the questions. Formally, \qrmodelname estimates the probability $p(R = 1|t,q)$, where $R$ is a binary random variable indicating whether the question $q$ should be retrieved ($R=1$) or not ($R=0$). $t$ and $q$ denote the query (topic) and the candidate clarifying question, respectively. The question relevance probability in the \qrmodelname model is estimated as follows:
\vspace{-0.15cm}
\begin{equation}
    p(R = 1|t,q) = \psi\big(\phi_T(t), \phi_Q(q)\big)~,
    \vspace{-0.1cm}
\end{equation}
where $\phi_T$ and $\phi_Q$ denote topic representation and question representation, respectively. $\psi$ is the matching component that takes the aforementioned representations and produces a question retrieval score. There are various ways to implement any of these components. 

We implement $\phi_T$ and $\phi_Q$ similarly using a function that maps a sequence of words to a $d$-dimensional representation ($V^s \rightarrow \mathbb{R}^d$). We use the BERT~\cite{devlin2018bert} model to learn these representation functions. BERT is a deep neural network with 12 layers that uses an attention-based network called Transformers~\cite{DBLP:journals/corr/VaswaniSPUJGKP17}. We initialize the BERT parameters with the model that is pre-trained for the language modeling task on Wikipedia and fine-tune the parameters on \dataname with 3 epochs. BERT has recently outperformed state-of-the-art models in a number of language understanding and retrieval tasks~\cite{devlin2018bert,Padigela:2019}. We particularly use BERT in our model to incorporate the knowledge from the vast amount of unlabeled data while learning the representation of queries and questions. In addition, BERT shows promising results in modeling short texts. 

The component $\psi$ is modeled using a fully-connected feed-forward network with the output
dimensionality of 2. Rectified linear unit (ReLU) is employed as the activation function in the hidden layers, and a softmax function is applied on the output layer to compute the probability of each label (i.e., relevant or non-relevant). To train \qrmodelname, we use a cross-entropy loss function.

\subsection{Question Selection Model}
\label{sec:qs}

In this section, we introduce a \textbf{Neu}ral \textbf{Q}uestion \textbf{S}election Model (\modelname) which selects questions with a focus on maximizing the precision at the top of the ranked list.
The main challenge in the question selection task is to predict whether a question has diverged from the query and conversation context. In cases where a user has given a negative answer(s) to previous question(s), the model needs to diverge from the history. In contrast, in cases where the answer to the previous question(s) is positive, questions on the same topic that ask for more details are preferred.
For example, as we saw in Figure~\ref{fig:exmaple}, when Robin answers the first question positively (i.e., being interested in dinosaur books), the second question tries to narrow down the information to a specific type of dinosaur. 

\modelname incorporates multiple sources of information. In particular, it learns from the similarity of a query, a question and the context as well as retrieval and performance prediction signals. In particular, \modelname outputs a relevance score for a given query $t$, question $q$, and conversation context $h$. Formally, \modelname can be defined as follows:
\vspace{-0.15cm}
\begin{equation}
    score = \gamma\big(\phi_T(t), \phi_H(\mathbf{h}), \phi_Q(q), \eta(t,\mathbf{h},q), \sigma(t,\mathbf{h},q)\big)~,
    \vspace{-0.1cm}
\end{equation}
where $\gamma$ is a scoring function for a given query representation $\phi_T(t)$, context representation $\phi_H(\mathbf{h})$, question representation $\phi_Q(q)$, retrieval representation $\eta(t,\mathbf{h},q)$, and query performance representation $\sigma(t,\mathbf{h},q)$. Various strategies can be employed to model each of the components of \modelname.

We model the components $\phi_T$ and $\phi_Q$ similarly to Section~\ref{sec:qr}. Further, the context representation component $\phi_H$ is implemented as follows:
\vspace{-0.15cm}
\begin{equation}
    \phi_H(\mathbf{h}) = \frac{1}{|\mathbf{h}|}\sum_{i}^{|\mathbf{h}|} \phi_{QA}(q_i,a_i)~,
    \vspace{-0.1cm}
\end{equation}
where $\phi_{QA}(q,a)$ is an embedding function of a question $q$ and answer $a$.
Moreover, the retrieval representation $\eta(t,\mathbf{h},q) \in \mathbb{R}^k$ is implemented by interpolating the retrieval score of the query, context and question (see Section~\ref{sec:docret}) and the score of the top $k$ retrieved documents is used. Finally, the query performance prediction (QPP) representation component $\sigma(t,\mathbf{h},q) \in \mathbb{R}^k$ consists of the performance prediction score of the ranked documents at different ranking positions (for a maximum of $k$ ranked documents). We employed the $\sigma$ QPP model for this component~\cite{DBLP:conf/spire/Perez-IglesiasA10}. We take the representations from the \texttt{[CLS]} layer of the pre-trained uncased BERT-Base model (i.e., 12-layer, 768-hidden, 12-heads, 110M parameters).
To model the function $\gamma$ we concatenate and feed  $\phi_T(t)$, $\phi_H(\mathbf{h})$, $\phi_Q(q)$, $\eta(t,\mathbf{h},q)$, and $\sigma(t,\mathbf{h},q)$ into a fully-connected feed-forward network with two hidden layers. We use ReLU as the activation function in the hidden layers of the network. We use a pointwise learning setting using a cross-entropy loss function.

\subsection{Document Retrieval Model}
\label{sec:docret}
Here, we describe the model that we use to retrieve documents given a query, conversation context, and current clarifying question as well as user's answer. We use the KL-divergence retrieval model~\cite{Lafferty:2001:DLM:383952.383970} based on the language modeling framework~\cite{DBLP:conf/sigir/PonteC98} with Dirichlet prior smoothing~\cite{Zhai:2017:SSM:3130348.3130377} where we linearly interpolate two likelihood models: one based on the original query, and one based on the questions and their respective answers.

For every term $w$ of the original query $t$, conversation context $\mathbf{h}$, the current question $q$, and answer $a$, the interpolated query probability is computed as follows:
\vspace{-0.1cm}
\begin{equation}
    \label{eq:ql}
    p(w| t, \mathbf{h}, q, a) = \alpha \times p(w|\theta_t) + (1-\alpha) \times p(w|\theta_{\mathbf{h},q,a})~,
    \vspace{-0.05cm}
\end{equation}
where $\theta_t$ denotes the language model of the original query, and $\theta_{\mathbf{h},q,a}$ denotes the language model of all questions and answers that have been exchanged in the conversation. $\alpha$ determines the weight of the original query and is tuned on the development set.

Then, the score of document $d$ is calculated as follows:
\vspace{-0.1cm}
\begin{equation}
    p(d|t, \mathbf{h}, q, a) = \sum_{w_k \in \tau} p(w_k| t, \mathbf{h}, q, a)\log(p(w_k|d)~,
    \vspace{-0.05cm}
\end{equation}
where $\tau$ is the set of all the terms present in the conversation. We use Dirichlet's smoothing for terms that do not appear in $d$. We use the document retrieval model for two purposes: \begin{inlinelist}
    \item ranking documents after the user answers a clarifying question;
    \item ranking documents of a candidate question as part of the \modelname (see Section~\ref{sec:qs}).
\end{inlinelist} Hence, the model does not see the answer in the latter case.

\vspace{-0.0cm}
\section{Experiments}
\label{sec:exp}

\subsection{Experimental Setup}

\partitle{Dataset.} 
We evaluate \qrmodelname and \modelname on \dataname, following a 5-fold cross-validation. We follow two strategies to split the data,
\begin{inlinelist}
\item \textbf{\dataname-T:} we split the train/validation/test sets based on topics. In this case, the model has not seen the test topics in the training data;
\item \textbf{\dataname-F:} here we split the data based on their facets. Thus, the same test topic might appear in the training set, but with a different facet.
\end{inlinelist}

In order to study the effect of multi-turn conversations with clarifying questions, we expand \dataname to include multiple artificially generated conversation turns. To do so, for each instance, we consider all possible combinations of questions to be asked as the context of conversation. Take $t_1$ as an example where we select a new question after asking the user two questions. Assuming that $t_1$ has four questions, all possible combinations of questions in the conversation context would be: $(q_1, q_2), (q_1, q_3), (q_1, q_4), (q_2, q_3), (q_2, q_4),$ $(q_3, q_4)$. Notice that the set of candidate clarifying questions for each multi-turn example would be the ones that have not appeared in the context. 
The number of instances grows significantly as we enlarge the length of the conversation, leading to a total of 907,366 instances in the collection. At each turn of the conversation, we select the question from all candidate questions of the same topic and facet, having the same conversation history. In other words, they share the same context. Since the total number of unique conversational contexts is 75,200, a model should select questions for 75,200 contexts from all 907,366 candidate questions.

\partitle{Question retrieval evaluation metrics.} 
We consider four metrics to evaluate the effectiveness of question retrieval models: mean average precision (MAP) and recall for the top 10, 20, and 30 retrieved questions (Recall@10, Recall@20, Recall@30). Our choice of measures is motivated by the importance of achieving high recall for this task.

\partitle{Question selection evaluation metrics.} 
Effectiveness is measured considering the performance of retrieval after adding the selected question to the retrieval model as well as the user answer.
Five standard evaluation metrics are considered: mean reciprocal rank (MRR), precision of the top 1 retrieved document (P@1), and normalized discounted cumulative gain for the top 1, 5, and 20 retrieved documents (nDCG@1, nDCG@5, nDCG@20). 
We use the relevance assessments as they were released by TREC. However, we modify them in such a way to evaluate the performance with respect to every facet. For instance, if one topic consists of 4 facets it is then broken into 4 different topics each inheriting its own relevance judgements from the TREC assessments.

The choice of evaluation metrics is motivated by considering three different aspects of the task. We choose MRR to evaluate the effect of asking clarifying questions on ranking the first relevant document. We report P@1 and nDCG@1 to measure the performance for scenarios where the system is able to return only one result. This is often the case with voice-only conversational systems. Moreover, we report nDCG@5 and nDCG@20 as conventional ranking metrics to measure the impact of asking clarifying questions in a traditional Web search setting. 
Notice that nDCG@20 is the preferred evaluation metric for the ClueWeb collection due to the shallow pooling performed for relevance assessments~\cite{DBLP:journals/ir/0002MC16, DBLP:conf/trec/ClarkeCSV11}.

\partitle{Statistical test.} We determine statistically significant differences using the two-tailed paired t-test with Bonferroni correction at a $99.9\%$ confidence interval ($p < 0.001$).

\begin{table}[]
    \centering
    \caption{Performance of question retrieval model. The superscript * denotes statistically significant differences compared to all the baselines ($p < 0.001$).}
     \vspace{-0.3cm}
    \begin{tabular}{l@{\quad}c@{\quad}c@{\quad}c@{\quad}c@{\quad}}
        \toprule
        \textbf{Method}     & MAP & Recall@10 & Recall@20 & Recall@30  \\
        \midrule
        QL                  & 0.6714 &  0.5917 &  0.6946 &  0.7076 \\
        BM25                & 0.6715 &  0.5938 &  0.6848 &  0.7076 \\
        RM3                 & 0.6858 &  0.5970 &  0.7091 &  0.7244 \\
        LambdaMART          & 0.7218 &  0.6220 &  0.7234 &  0.7336  \\
        RankNet             & 0.7304 &  0.6233 &  0.7314 &  0.7500  \\
        \qrmodelname             & \textbf{0.8349*} & \textbf{0.6775*} & \textbf{0.8310*} &\textbf{ 0.8630*} \\
        \bottomrule
    \end{tabular}
    \label{tab:qret}
     \vspace{-0.4cm}
\end{table}

\begin{table*}
    \centering
    \caption{Performance comparison with baselines.
    \textit{WorstQuestion} and \textit{BestQuestion} respectively determine the lower and upper bounds.
    The superscript * denotes statistically significant differences compared to all the baselines ($p < 0.001$).
    }
    \vspace{-0.3cm}
    \label{tab:results}
    \begin{tabular}{p{2cm}c@{\quad}c@{\quad}c@{\quad}c@{\quad}c@{\quad}c@{\quad}cc@{\quad}c@{\quad}c@{\quad}c@{\quad}c@{\quad}c@{\quad}c}
    \toprule
     \multirow{2}{*}{\textbf{Method}} & \multicolumn{5}{c}{\textbf{\dataname-T Dataset}}  & \multicolumn{5}{c}{\textbf{\dataname-F Dataset}} \\
     \cmidrule{2-6} \cmidrule{8-14} 
    
     & MRR & P@1 & nDCG@1 & nDCG@5 & nDCG@20 && MRR & P@1 & nDCG@1 & nDCG@5 & nDCG@20 \\
    \midrule
    OriginalQuery & 0.2715 & 0.1842 & 0.1381 & 0.1451 & 0.1470 && 0.2715 & 0.1842 & 0.1381 & 0.1451 & 0.1470 \\
    $\sigma$-QPP & 0.3570 &  0.2548 &  0.1960 &  0.1938 &  0.1812 && 0.3570 &  0.2548 &  0.1960 &  0.1938 &  0.1812\\
    LambdaMART & 0.3558 &  0.2537 &  0.1945 &  0.1940 &  0.1796 && 0.3501 &  0.2478 &  0.1911 &  0.1896 &  0.1773\\
    RankNet & 0.3573 &  0.2562 &  0.1979 &  0.1943 &  0.1804 && 0.3568 &  0.2559 &  0.1986 &  0.1944 &  0.1809\\
    \modelname & \textbf{0.3625*} &  \textbf{0.2664*} &  \textbf{0.2064*} & \textbf{0.2013*} &  \textbf{0.1862*} & & \textbf{0.3641*} &  \textbf{0.2682*} &  \textbf{0.2110*} &  \textbf{0.2018*} &  \textbf{0.1867*} \\
    \midrule \midrule
    WorstQuestion 
    & 0.2479 &  0.1451 &  0.1075 &  0.1402 &  0.1483 && 0.2479 &  0.1451 &  0.1075 &  0.1402 &  0.1483\\
    BestQuestion
    & 0.4673 &  0.3815 &  0.3031 &  0.2410 &  0.2077 && 0.4673 &  0.3815 &  0.3031 &  0.2410 &  0.2077 \\
    \bottomrule
    \end{tabular}
\end{table*}

\partitle{Compared methods.} 
We compare the performance of our question retrieval and selection models with the following methods:
\newpage
\begin{itemize}[leftmargin=*]
    \item {Question retrieval:} 
        \begin{itemize}
            \item \textit{BM25, RM3, QL:} we index all the questions using Galago.\footnote{\url{https://sourceforge.net/p/lemur/galago/}} Then, for a given query we retrieve the documents using BM25~\cite{DBLP:conf/trec/RobertsonWJHG94}, RM3~\cite{DBLP:conf/sigir/LavrenkoC01}, and QL~\cite{DBLP:conf/sigir/PonteC98} models.
            \item \textit{LambdaMART, RankNet:} for every query-question pair, we use the scores obtained by BM25, RM3, and QL as features to train LambdaMART~\cite{DBLP:journals/ir/WuBSG10} and RankNet~\cite{DBLP:conf/icml/BurgesSRLDHH05} implemented in RankLib.\footnote{\url{https://sourceforge.net/p/lemur/wiki/RankLib/}} For every query, we consider all irrelevant questions as negative samples.
        \end{itemize}
    \item {Question selection:}
        \begin{itemize}[leftmargin=*]
            \item \textit{OriginalQuery} reports the performance of the document retrieval model only with the original query (Eq. \eqref{eq:ql} with $\alpha=1$).
            \item \textit{$\sigma$-QPP:} we use a simple yet effective query performance predictor, $\sigma$~\cite{DBLP:conf/spire/Perez-IglesiasA10} as an estimation of a question's quality. We calculate the $\sigma$ predictor of the document retrieval model with the following input: original query, the context, and candidate questions. We then select the question with the highest $\sigma$ value.
            \item \textit{LambdaMART, RankNet:} we consider the task of question selection as a ranking problem where a list of candidate questions should be ranked and the one with the highest rank is chosen. Therefore, we use LambdaMART~\cite{DBLP:journals/ir/WuBSG10} and RankNet~\cite{DBLP:conf/icml/BurgesSRLDHH05} as two LTR baselines. The list of features are: \begin{inlinelist}
                \item a flag determining if a question is open or not;
                \item a flag indicating if the answer to the last question in the context is yes or no;
                \item $\sigma$~\cite{DBLP:conf/spire/Perez-IglesiasA10} performance predictor of the current question;
                \item the Kendall's $\tau$ correlation of the ranked list at 10 and 50 of the original query and the current question;
                \item the Kendall's $\tau$ correlation of the ranked list at 20 and 50 of the current question and previous question-answer pairs in the context;
                \item Similarity of the current question and the query based on their BERT representations;
                \item Similarity of the current question and previous question-answer pairs in the context based on their BERT representations.
                \end{inlinelist}
            \item \textit{BestQuestion, WorstQuestion:} in addition to all the baselines, we also report the retrieval performance when the worst and the best question is selected for an instance. BestQuestion (WorstQuestion) selects the candidate question for which the MRR value of the retrieval model is the maximum (minimum). Note that the retrieval scores are calculated knowing the selected question and its answer (i.e., oracle model). Our goal is to show the upper and lower bounds.
        \end{itemize}
\end{itemize}

\begin{figure}
    \centering
    \includegraphics[trim=7 7 7 7, clip,width=\columnwidth]{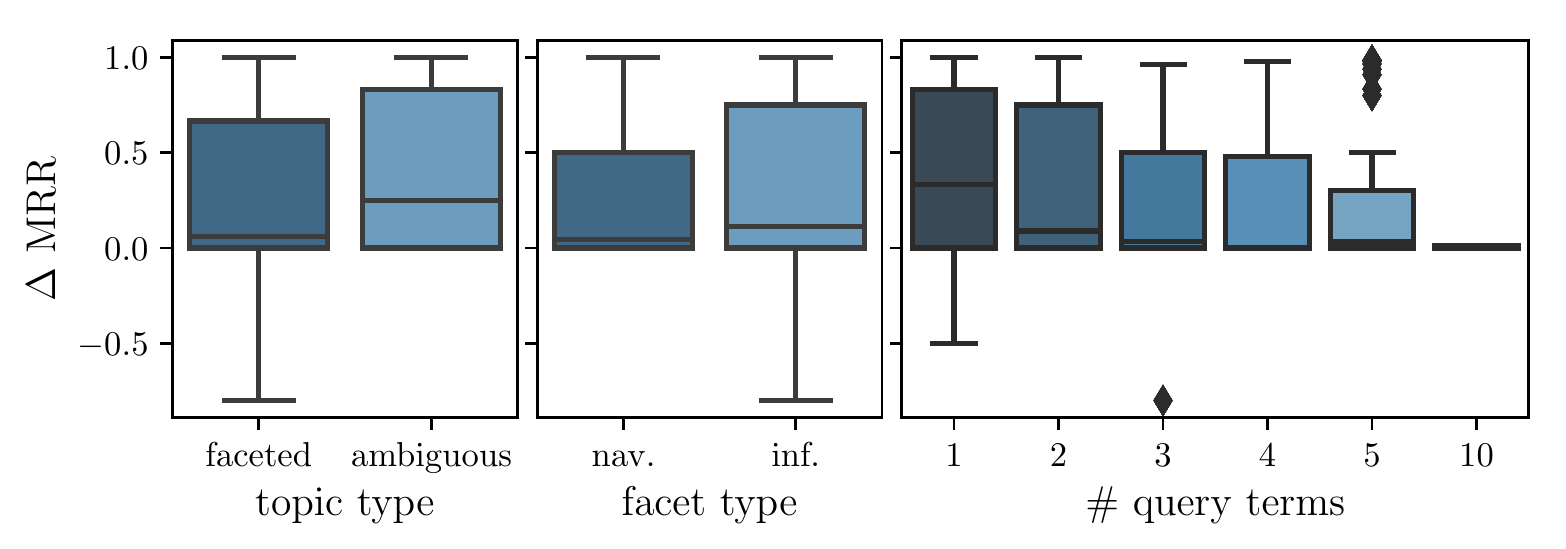}
     \vspace{-0.8cm}
    \caption{Impact of topic type, facet type, and query length on the performance of BestQuestion oracle model, compared to OriginalQuery.}
    \label{fig:oracle}
\end{figure}

\begin{figure*}
    \centering
    \vspace{-0.5cm}
    \begin{subfigure}[t]{0.5\textwidth}
        ~~~~~
        \includegraphics[trim=53 228 5 6, clip,width=\columnwidth]{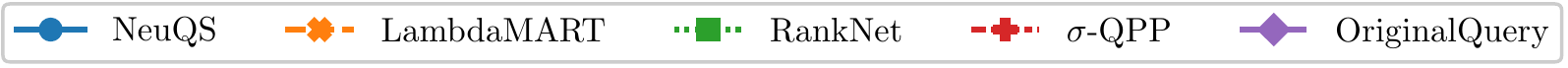}
    \end{subfigure}
    
    \begin{subfigure}[t]{0.32\textwidth}
        \includegraphics[trim=7 7 7 7, clip,width=\columnwidth]{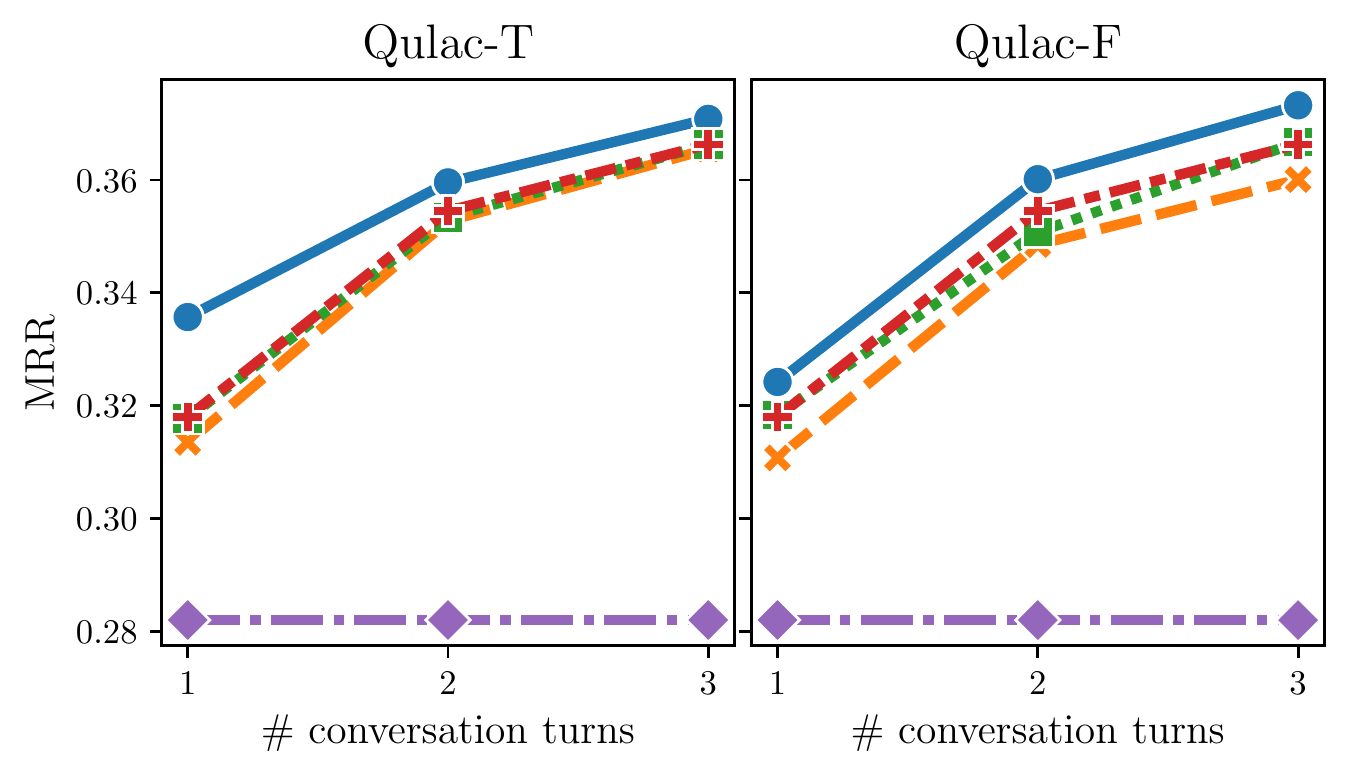}
    \end{subfigure}
    ~
    \begin{subfigure}[t]{0.32\textwidth}
        \includegraphics[trim=7 7 7 7, clip,width=\columnwidth]{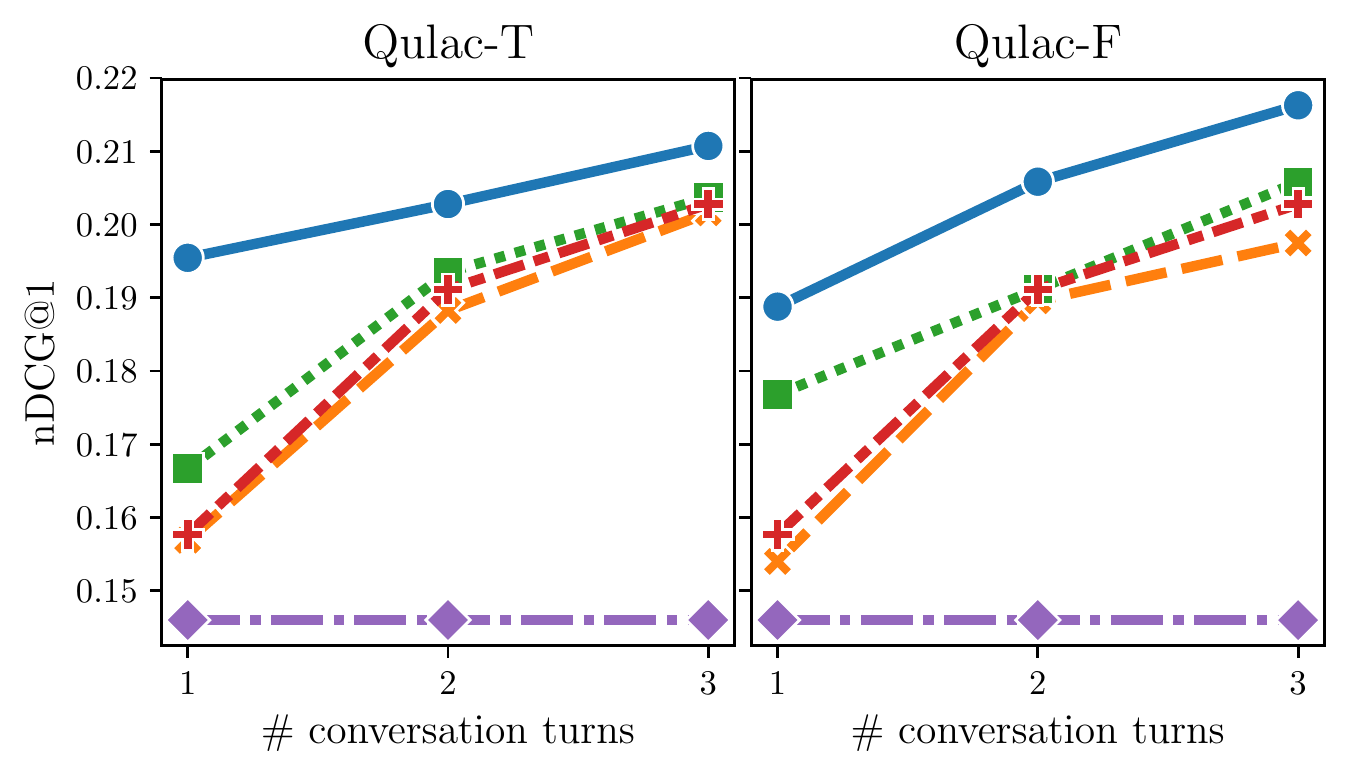}
    \end{subfigure}
    ~
    \begin{subfigure}[t]{0.32\textwidth}
        \includegraphics[trim=7 7 7 7, clip,width=\columnwidth]{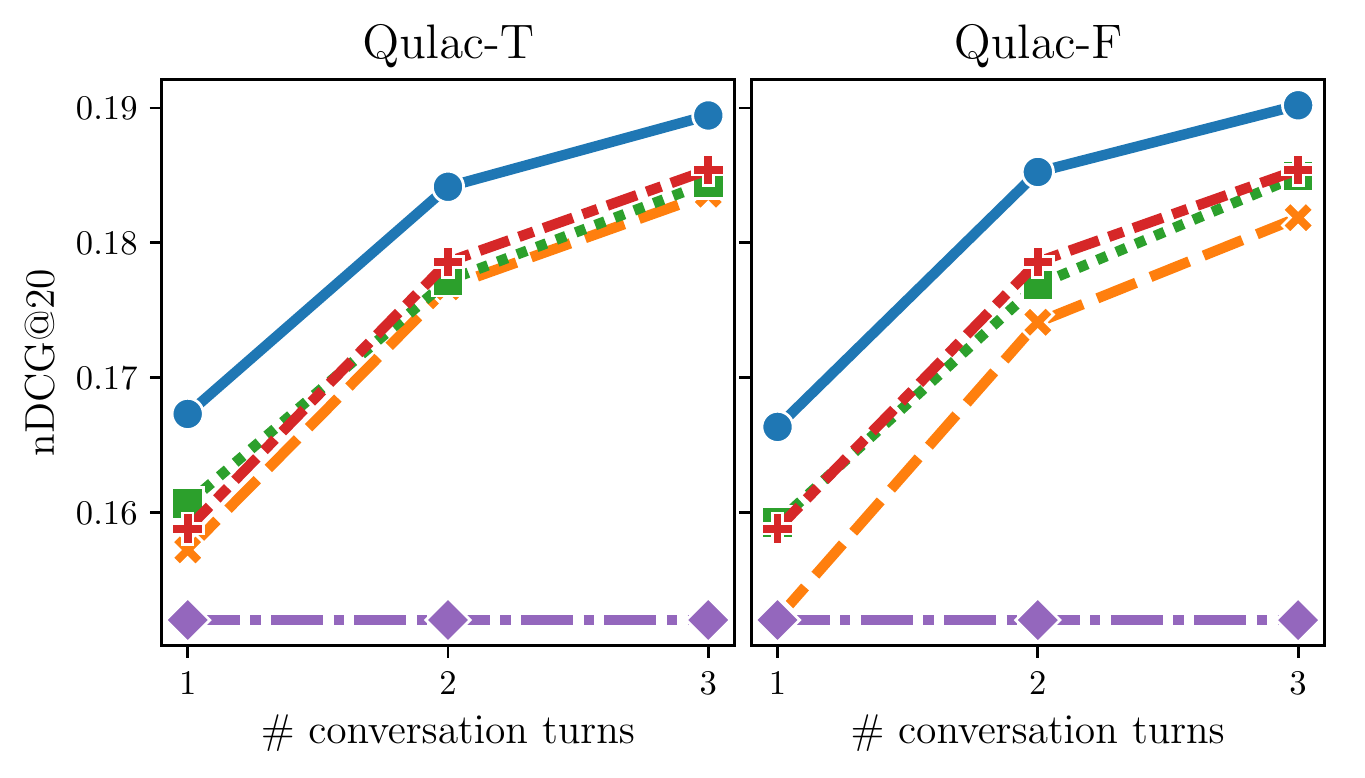}
    \end{subfigure}

    \vspace{-0.3cm}
    \caption{Performance comparison with the baselines for different number of conversation turns ($k \in \{1,2,3\}$).}
    \label{fig:turns}
    \vspace{+0.1cm}
\end{figure*}

\vspace{-0.3cm}
\subsection{Results and Discussion}
\partitle{Question retrieval.}
Table \ref{tab:qret} shows the results of question retrieval for all the topics. As we see, \qrmodelname is able to outperform all baselines. It is worth noting that the model's performance gets better as the number of retrieved documents increases. This indicates that \qrmodelname is able to capture the relevance of query and questions when they lack common terms. In fact, we see that all term-matching retrieval models such as BM25 are significantly outperformed in terms of all evaluation metrics.

\partitle{Oracle question selection: performance.}
Here we study the performance of an oracle model, i.e. assuming that an oracle model is aware of the answers to the questions. The goal is to show to what extent clarifying questions can improve the performance of a retrieval system.
As we see in the lower rows of Table \ref{tab:results}
selecting best questions (BestQuestion model) helps the model to achieve substantial improvement, even in the case that the retrieval model is very simple. This shows the high potential gain of asking good clarifying questions on the performance of a conversational system. Particularly, we examine the relative improvement of the system after asking only one question and observe that BestQuestion achieves over 100\% relative improvement in terms of different evaluation metrics
(MRR: $0.2820 \rightarrow 0.5677$, P@1: $0.1933 \rightarrow 0.4986$, nDCG@1: $0.1460 \rightarrow 0.3988$, nDCG@5: $0.1503 \rightarrow 0.2793$, nDCG@20: $0.1520 \rightarrow 0.2265$).
It is worth mentioning that we observe the highest relative improvements in terms of nDCG@1 (=173\%) and P@1 (=158\%), exhibiting a high potential impact on voice-only conversational systems.

\partitle{Oracle question selection: impact of topic type and length.}
We analyze the performance of BestQuestion based on the number of query terms and topic type. We see that the relative improvement of BestQuestion is negatively correlated with the number of query terms (Pearson's $r=-0.2$, $p \ll 0.001$), suggesting that shorter queries require clarification in more cases. Also, comparing the topic types (ambiguous vs. faceted), we see a significant difference in the relative improvement. The average $\Delta$MRR for ambiguous topics is 0.3858, compared with the faceted topics with average $\Delta$MRR of 0.2898. The difference was statistically significant (2-way ANOVA, $p \ll 0.001$).

\partitle{Question selection.}
Table \ref{tab:results} presents the results of the document retrieval model taking into account a selected question together with its answer.
We see that all models outperform OriginalQuery, confirming that asking clarifying questions is crucial in a conversation, leading to high performance gain. 
For instance, compared to OriginalQuery, a model as simple as $\sigma$-QPP achieves a 31\% relative improvement in terms of MRR. 
Also, \modelname consistently outperforms all the baselines in terms of all evaluation metrics on both data splits. All the improvements are statistically significant. Moreover, \modelname achieves a remarkable improvement in terms of both P@1 and nDCG@1. These two evaluation metrics are particularly important for voice-only conversational systems where the system must return only one result to the user. The obtained improvements highlight the necessity and effectiveness of asking clarifying questions in a conversational search system, where they are perceived as natural means of interactions with users.

\partitle{Impact of data splits.} We compare the performance of models on both \dataname-T and \dataname-F data splits. We see that the LTR baselines perform worse on \dataname-F. Notice that the performance difference of LambdaMART among the splits is statistically significant in terms of all evaluation metrics ($p < 0.001)$. RankNet, on the other hand, exhibits a more robust performance, i.e., the difference of its performance on the two splits is not statistically significant. Unlike the baselines, \modelname exhibits a significant improvement in terms of all evaluation metrics on \dataname-F ($p < 0.05)$, except for nDCG@5. This suggests that the baseline models are prone to overfitting on queries and conversations in the training data. As mentioned, \dataname-F's train and test sets may have some queries and questions in common, hurting models that are weak at generalization.

\partitle{Impact of number of conversation turns.} Figure~\ref{fig:turns} shows the performance of \modelname as well as the baselines for different conversation turns. We evaluate different models at $k$ turns ($k \in \{1,2,3\}$). We see that the performance of all models improves as the conversation advances to multiple turns. Also, we see that all the models consistently outperform the OriginalQuery baseline at different number of turns. Finally, we see that \modelname exhibits robust performance, outperforming all the baselines at different turns.

\partitle{Impact of clarifying questions on facets.}
We study the difference of MRR between \modelname and OriginalQuery on all facets. Note that for every facet we average the performance of \modelname at different conversation turns. Our goal is to see how many facets are impacted positively by asking clarifying questions.
\modelname is improves the effectiveness of retrieval by selecting relevant questions for a considerable number of facets on both data splits. In particular, the performance for 45\% of the facets is improved by asking clarifying questions, whereas the performance for 19\% is worse.

\begin{table*}[]
    \centering
    \caption{Failure and success examples of \modelname. Failure and success are measured by the difference in performance of \modelname and OriginalQuery in terms of MRR ($\Delta$MRR).}
    \vspace{-0.3cm}
    \begin{adjustbox}{max width=\textwidth}
        \begin{tabular}{p{2cm}p{4.5cm}p{4.5cm}p{4.5cm}c}
            \toprule
             \textbf{Query} & \textbf{Facet Description} & \textbf{Selected Question} & \textbf{User's Answer}&\textbf{$\Delta$MRR}\\
            \midrule
            dog heat & What is the effect of excessive heat on dogs? & Would you like to know how to care for your dog during heat? & No, I want to know what happens when a dog is too hot. & $-$0.86\\
            \midrule
            sit and reach test & How is the sit and reach test properly done? & Do you want to know how to perform this test? & Yes, I do. & $-$0.75\\
            \midrule
            alexian brothers hospital & Find Alexian Brothers hospitals. & Are you looking for our schedule of classes or events? & No, I don't need that. & $-$0.54\\
            \midrule \midrule
            east ridge high school & Information about the sports program at East Ridge High School in Clermont, Florida & What information about East Ridge High School are you looking for? & I'm looking for information about their sports program. & $+$0.96\\
            \midrule
            euclid & Find information on the Greek mathematician Euclid. & Do you want a biography? & Yes. & $+$0.93\\
            \midrule
            rocky mountain news & Who are the sports reporters for the Rocky Mountain News? & Would you like to read recent news about the Rocky Mountain News? & No, I just want a list of the reporters who write the sports for the Rocky Mountain News. & $+$0.88\\
            \bottomrule
        \end{tabular}
    \end{adjustbox}
    \label{tab:failure}
        \vspace{-0.3cm}
\end{table*}

\partitle{Case study: failure and success analysis.} 
Finally, we analyze representative cases of failure and success of our proposed framework. We list three cases where selecting questions using \modelname improves the retrieval performance, as well as three other examples in which the selected questions lead to decreased performance. $\Delta$MRR reports the difference of the performance of \modelname and OriginalQuery in terms of MRR. As we see, the first three examples show the selected questions that hurt the performance (i.e., $\Delta$MRR < 0.0). The first row is an example where the user's response to the question is negative; however, the user provides additional information about their information need (i.e., facet). We see that even though the user has provided additional information, the performance drops. This is perhaps due to existence of no common terms between the additional information (i.e., ``dog is too hot'') and the facet (i.e., ``\textit{excessive} \textit{heat} on dogs''). 
This is more evident when we compare this example with a successful answer: ``No, I would like to know the effects of \textit{excessive} \textit{heat} on dogs.'' The second row of the table shows a case where the answer to the question is positive, but there is no common terms between the question and the facet. Again, the intuition here is that the retrieval model is not able to take advantage of additional information when it has no terms in common with relevant documents. The third row of the table shows another failure example where the selected question is not relevant to the facet and the user provides no additional information. This is a typical failure example where the system does not get any positive feedback, but could still use the negative feedback to improve the ranking. This can be done by diverging from the documents that are similar to the negative feedback.

As for the success examples, we have listed three types. The first example (``east ridge high school'') is where the system is able to ask an open question. Open questions are very hard to formulate for open-domain information-seeking scenarios; however, it is more likely to get useful feedback from users in response to such questions. The fifth row shows an example of a positive feedback. The performance gain, in this case, is perhaps due to the existence of term ``biography'' in the question which would match with relevant documents. 
It is worth noting that the question and the query in this example have no common terms. This highlights the importance of employing a language-representation-based question retrieval model (e.g., \qrmodelname) as opposed to term-matching IR models.
The last example shows a case where the answer is negative, but the user is engaged in the conversation and provides additional information about the facet. We see that the answer contains keywords of the facet description (i.e., ``reporters,'' ``sports''), improving the score of relevant documents that contain those terms. 

\section{Limitations and Future Work}

Every data collection comes with some limitations. The same is valid for \dataname. First, the dataset was not collected from actual conversations. This decision was mainly due to the unbalanced workload of the two conversation participants. In our crowdsourcing HITs, the task of question generation required nearly 10 times more effort compared to the task of question answering. This makes it challenging and more expensive to pair two workers as participants of the same conversation. There are some examples of this approach in the literature~\cite{DBLP:conf/emnlp/ChoiHIYYCLZ18,DBLP:journals/corr/abs-1808-07042}; however, they address the task of reading comprehension, a task that is considerably simpler than identifying topic facets. A possible future direction is to provide a limited number of pre-generated questions (say 10) to the workers to select from, so that the complexity of the task would be significantly reduced.

Furthermore, \dataname is built for single-turn conversations (i.e., one question; one answer). Even though there are questions that can be asked after one another to form a multi-turn conversation, our data collection approach does not guarantee the existence of multi-turn conversations that involve the same participants. Also, we believe that the quality of generated clarifying questions highly depends on how well the selected commercial search engine is able to diversify the result list. We aimed to minimize this bias by asking workers to scan at least three pages of the result list. 
Also, the questions added by expert annotators guarantees the coverage of all facets (see Section~\ref{sec:verification}).
Finally, as we mentioned, faceted and ambiguous queries are good examples of topics that a conversational system needs to clarify; however, this task cannot be limited only to such queries. One can collect a similar data for exploratory search scenarios, where asking questions can potentially lead to more user engagement while doing exploratory search.

In this work, our main focus was on question selection. There are various directions that can be explored in the future. One interesting problem is to explore various strategies of improving the performance of the document retrieval model as new information is added to the model. Moreover, we assumed the number of conversation turns to be fixed. Another interesting future direction is to model the system's confidence at every stage of the conversation so that the model is able to decide when to stop asking questions and present the result(s).

\section{Conclusions}
In this work, we introduced the task of asking clarifying questions in open-domain information-seeking conversations. We proposed an evaluation methodology which enables offline evaluation of conversational systems with clarifying questions. Also, we constructed and released a new data collection called \dataname, consisting of 762 topic-facet pairs with over 10K question-answer pairs. We further presented a neural question selection model called \modelname along with models on question and document retrieval. \modelname was able to outperform the LTR baselines significantly. The experimental analysis provided many insights of the task. In particular, experiments on the oracle model demonstrated that asking only one good clarifying question leads to over 150\% relative improvement in terms of P@1 and nDCG@1. Moreover, we observed that asking clarifying questions improves the model's performance for a substantial percentage of the facets. In some failure cases, we saw that a more effective document retrieval model can potentially improve the performance. 
Finally, we showed that, asking more clarifying questions leads to better results, once again confirming the effectiveness of asking clarifying questions in a conversational search system.

\newpage
\begin{acks}
This work was supported in part by the RelMobIR project of the \grantsponsor{}{Swiss National Science Foundation (SNSF)}{http://www.snf.ch/en/Pages/default.aspx}, in part by the Center for Intelligent Information Retrieval, and in part by NSF grant IIS-1715095. Any opinions, findings and conclusions or recommendations expressed in this material are those of the authors and do not necessarily reflect
those of the sponsors.

\end{acks}

\bibliographystyle{ACM-Reference-Format}
\bibliography{main} 

\end{document}